\documentclass[letterpaper, 10 pt, conference]{ieeeconf} 

\IEEEoverridecommandlockouts
\usepackage{tabularx, graphicx,
  amsmath,amsfonts,siunitx,pifont,amssymb}
\usepackage[table]{xcolor}
\usepackage{xcolor, bm, float, multirow, multicol, courier, mathtools, url}
\usepackage[ruled,vlined]{algorithm2e} \pdfminorversion=4
\let\labelindent\relax \usepackage{enumitem} \usepackage{graphicx}
\usepackage{adjustbox}

\usepackage{pdfpages}
\usepackage{cite}
\usepackage{balance}
\usepackage{hyperref}

\def\secref#1{Sec.~\ref{#1}}
\def\figref#1{Fig.~\ref{#1}}
\def\tabref#1{Table~\ref{#1}}
\def\eqref#1{Eq.~(\ref{#1})}

\title{\LARGE \bf{Actively Mapping Industrial Structures\\ with Information Gain-Based Planning on a Quadruped Robot}}

\author{Yiduo Wang, Milad Ramezani and Maurice Fallon
	\thanks{This research is supported by the ESPRC ORCA Robotics Hub (EP/R026173/1). M. Fallon is supported by a Royal Society University Research Fellowship.
	}
	\thanks{The authors are with the Oxford Robotics Institute, University of Oxford, UK.
		{\tt\small \{ywang, milad, mfallon\}@robots.ox.ac.uk}}%
}


\begin{document}
	
\setlength{\abovedisplayskip}{4pt}
\setlength{\belowdisplayskip}{4pt}
	
\maketitle 
\thispagestyle{empty} 
\pagestyle{empty}
	

\begin{abstract}
In this paper, we develop an online active mapping system to enable a quadruped robot
to autonomously survey large physical structures. 
We describe the perception, planning and control modules needed to scan 
and reconstruct an object of interest, without requiring a prior model. The system builds a  
voxel representation of the object, and iteratively determines the Next-Best-View (NBV) to extend the representation, according to both the reconstruction itself and to avoid collisions with the environment. By computing the expected
information gain of a set of candidate scan locations sampled on the as-sensed terrain map, 
as well as the cost of reaching these candidates, the 
robot decides the 
NBV for further exploration. The robot plans an optimal path 
towards the NBV, avoiding obstacles and un-traversable terrain. Experimental results on both 
simulated and real-world environments show the capability and efficiency of our system. Finally
we present a full system demonstration on the real robot, the ANYbotics ANYmal, 
autonomously reconstructing a building facade and an industrial structure.  

\end{abstract}
	
\section{Introduction}

In the context of robotics, active perceptual planning refers to exploration by
a mobile robot equipped with sensors to conduct a survey of an object or  
environment of interest. It can be of assistance for the regular inspection and monitoring of remote or dangerous facilities such as offshore platforms.

Although active mapping has been investigated for many applications such as 
inspection~\cite{Hollinger2013} and virtual modelling~\cite{Kriegel2015}, 
and on robotic platforms as varied as aerial~\cite{Bircher2018, Delmerico2018}, wheeled~\cite{Isler2016, Vasquez-Gomez2014a} 
and underwater robots~\cite{Ho2018, Franz2016}, the 
online deployment of such a system on a real robot is still a challenge, thus requiring 
further investigation. 

Advances in quadruped mobility and hardware reliability have been significant and the first industrial
prototypes are being tested on live industrial facilities \cite{gehring2019fsr}. Quadrupeds can cover the same terrain
as wheeled or tracked robots but can also cross mobility hazards and climb stairs. While UAVs are being actively
deployed for these kinds of missions, it is difficult to operate aerial platforms within confined
spaces and their sensing payload is limited.

\begin{figure}[t]
    \centering
     \vspace{0.3cm}
    \resizebox{\columnwidth}{!}{%
    \includegraphics[width=0.45\textwidth,trim={0cm 0cm 0cm 10cm},clip]{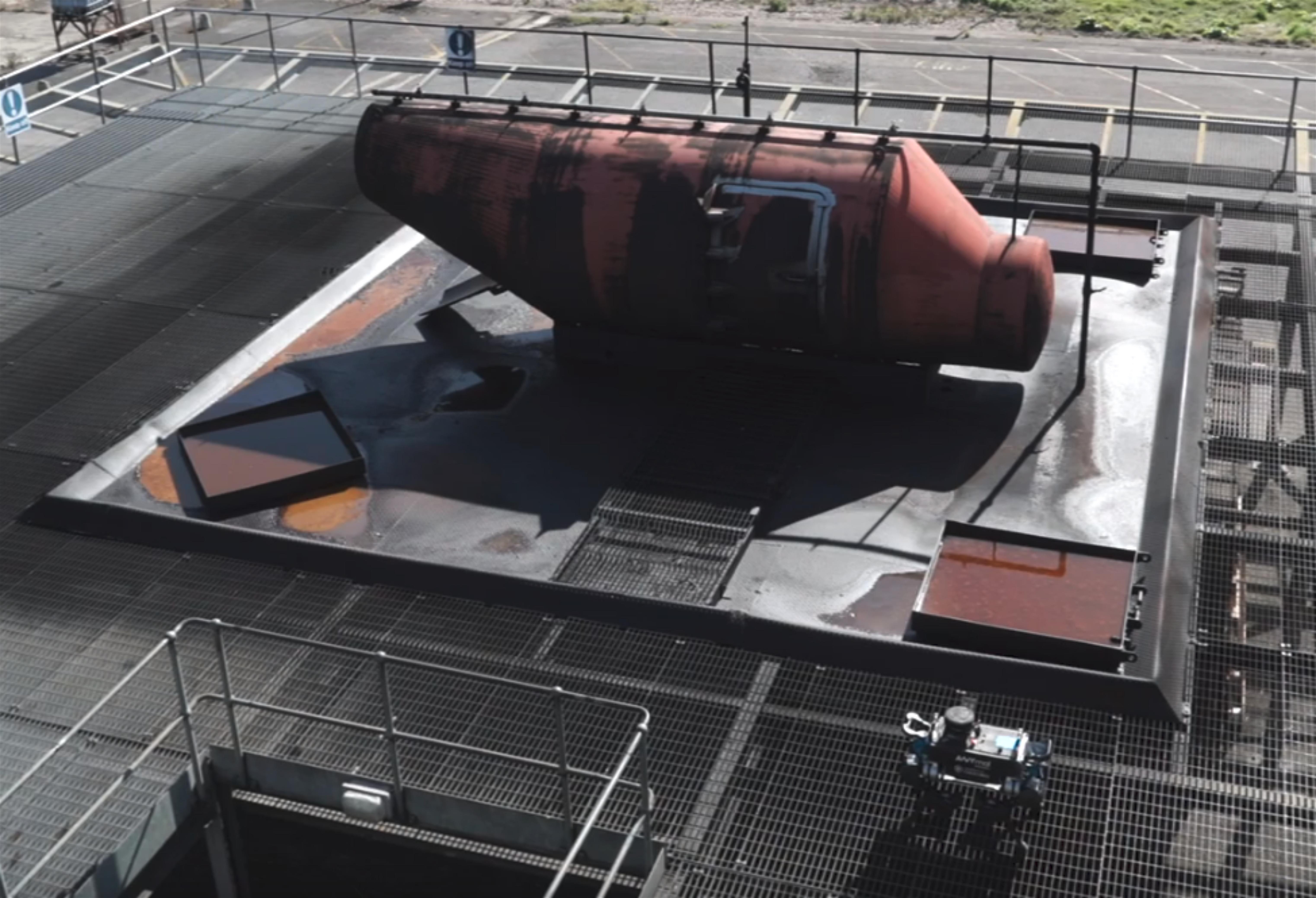}
    }
    \resizebox{\columnwidth}{!}{%
    \includegraphics[width=0.45\textwidth,trim={2cm 0cm 0cm 3cm},clip]{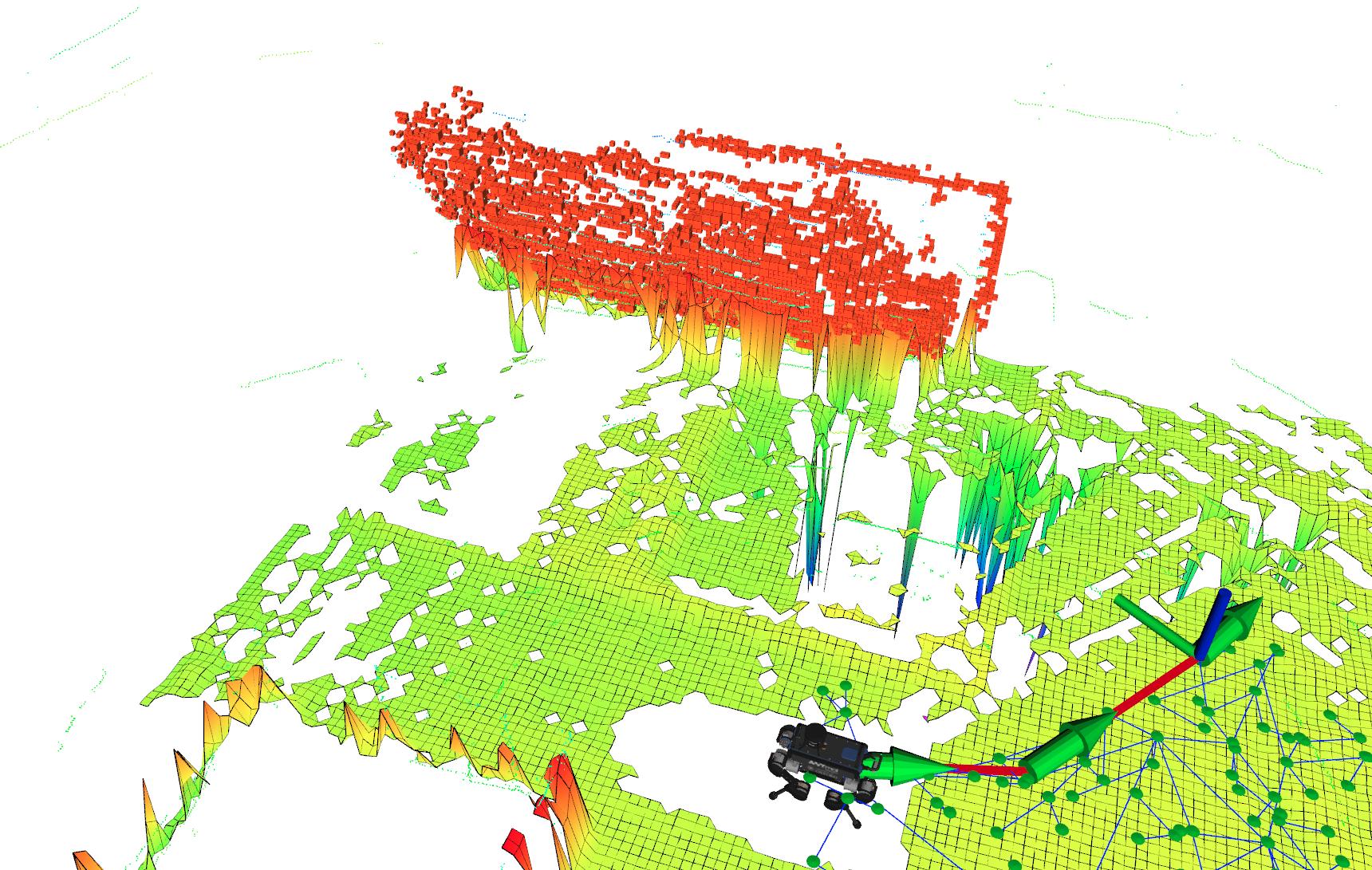}
    }
    \caption{Top: The ANYmal robot actively mapping a mockup helicopter at the Fire Service College in Gloucestershire, UK, as shown at \url{https://youtu.be/K348uuCB8gY}. The safety stanchions, stairwells and the skirt under the helicopter are mobility hazards which our approach can avoid. Bottom: the state of the mapping system showing the object reconstruction, LiDAR elevation map, an RRT plan and the next walking goal.}
    \label{fig:FscHelicopterDeck}
    \vspace{-2em}
\end{figure}

Our approach is to build and maintain an accurate 3D model of the object of interest as well as the local environment
and to use it to plan actions of a quadruped robot to improve and extend the model.

Our active mapping framework adapts the Information Gain (IG) approach originally formulated in~\cite{Isler2016}, 
which focused on the IG formulation with minimum attention to the problems that an actual robot faces in realistic inspections. 

We aim for a more realistic validation in an industrial setting (\figref{fig:FscHelicopterDeck}).
The object of interest will be of unknown shape, surrounded by uneven terrain and mobility hazards
and our solution will be embodied on the ANYmal quadruped robot. 
We present a complete active mapping system in the experimental section of this paper. 

The contributions of our research are as follows:

\begin{itemize}
    \item Implementation of an active mapping system on a quadruped, enabling the robot to not only traverse an unstructured environment, but also to scan an object of interest in an optimal manner using LiDAR. 
    \item Formulation of the approach as a Next-Best-View (NBV) problem, which determines the best pose for the robot to conduct the next scan on the basis of metrics drawn from the partial reconstruction (information gain) and the environment map (cost of mobility). 
    \item Evaluation of the system in simulated and real-world environments and the real-time deployment of our system on the ANYmal quadruped robot.
\end{itemize}

The remainder of this paper is organised as follows. In Section \ref{sec:RelatedWork} we discuss 
related works with our method detailed in Section \ref{sec:Method}. Experimental results 
are given in Section \ref{sec:Experiments} before discussing conclusions and future work in 
Section \ref{sec:FutureWork}.

\section{Related Works}
\label{sec:RelatedWork}
First we will review existing active mapping systems grouped according to two aspects: the prior assumptions made and the type of representation used.

\subsection{Prior Assumptions}

Active mapping systems are typically divided between model-based 
and model-free approaches~\cite{Karaszewski2016}. Model-based methods are essential for routine survey and 
inspection. They are typically applied in industrial scenarios because 
CAD models are often available~\cite{Chen2011}. 

Blaer and Allen~\cite{Blaer2007} designed a model-based 
system for 3D mapping of large environments by solving an Art Gallery Problem (AGP) and a Travelling
Salesman Problem (TSP) for path optimisation.
The system of Hollinger~\textit{et al.}~\cite{Hollinger2013} was designed to decrease the 
uncertainty in a ship hull surface mesh reconstruction and improve its quality. Their system assumed the 
availability of an initial mesh reconstruction and planned viewpoints that could inspect 
high uncertainty areas of the mesh. 

Model-free active vision systems are more versatile and can be applied to a wider variety of objects and sites. 
For instance, the system of Bircher~\textit{et al.}~\cite{Bircher2016, Bircher2018} aimed to explore unknown spaces of different scales, 
and the approach of Kriegel~\textit{et al.}~\cite{Kriegel2012, Kriegel2015} was designed to reconstruct objects of arbitrary shape but confined size. 

Model-based approaches can also be adapted to incorporate uncertainties in the prior model and to improve the quality of reconstruction. 
In a work following up on~\cite{Hollinger2013}, Hover~\textit{et al.}~\cite{Franz2016} addressed the potential lack of prior information by carrying out a coarse-to-fine multi-stage survey.

\subsection{Representations}

Many active mapping systems employ either surface mesh or voxel space representations. However, there are works such as~\cite{Kriegel2012, Kriegel2013} which benefited from both - surface mesh for reconstruction and voxel space for collision avoidance.

Hollinger~\textit{et al.}~\cite{Hollinger2013} and Hover~\textit{et al.}~\cite{Franz2016} utilised a surface mesh to precisely reconstruct ship hulls with the mesh providing information about surface coverage, boundaries and holes. 
Schmid~\textit{et al.}~\cite{schmid2012view} used a coarse Digital Surface Model (DSM) as a prior 
map to plan viewpoints for a UAV. Their algorithm's complexity is linearly proportional to the 
size of the site, limiting the applicability of this system to smaller scale environments.

In volumetric approaches, view selection metrics such as IG are normally used to determine an NBV. 

Bircher~\textit{et al.}~\cite{Bircher2016, Bircher2018} utilised a volumetric representation 
with their receding horizon planning strategy to progressively explore unknown environments. They grew a Rapidly-exploring Random Tree (RRT)~\cite{LaValle1998} and selected the best branch based upon IG that evaluates the amount of observable unknown space.

Isler~\textit{et al.}~\cite{Isler2016} proposed a collection of Volumetric Information (VI) 
measures. \textit{Occlusion Aware} computes the entropy of all observable voxels. 
\textit{Unobserved Voxel} only counts unknown voxels, thereby inclining the system towards exploring void spaces. \textit{Rear Side Voxel} and \textit{Rear Side Entropy} focus the sensor on the object of interest. \textit{Proximity Count} was shown to be advantageous in ensureing coverage of the object in their experiments but has a disadvantage of potentially pointing the sensor away from the object. The authors demonstrated these measures on a KUKA Youbot with a 5 Degree of Freedom (DoF) arm in an office room. 

Delmerico~\textit{et al.}~\cite{Delmerico2018} then conducted more 
experiments, comparing their VI formulations with the approaches of \cite{Kriegel2013} 
and \cite{Vasquez-Gomez2014}, to determine the best choice for the NBV 
selection in a volumetric representation. 

We aim for a model-free system able to fully scan an object of interest but also to explore in an
unknown environment, therefore our system minimises the entropy/uncertainty in the environment while focusing on the object. 
We pair this with an octree as our robot's volumetric representation, storing occupancy probability using OctoMap~\cite{OctoMap13}.

\subsection{Localization and Mapping}

Accurate mapping is crucial for precise model reconstruction and active planning. When a prior model is available, it can be used for pure localization, however a model-free approach requires a complete Simultaneous Localisation and Mapping (SLAM) system - itself an active research problem.

By its nature, a SLAM system will drift during exploration. Planning methods which use rigid representations such as a single OctoMap would struggle to respond to loop closures. The approach of~\cite{Ho2018} is interesting---using a deformable reconstruction with \textit{Virtual Occupancy Maps} attached to a pose graph to be flexible to new loop closures. For our real world experiments, we used a rigid map representation but are interested in this approach for future work.

\section{Method}
\label{sec:Method}

\begin{figure}[t]
    \centering
    \resizebox{\columnwidth}{!}{%
    \includegraphics[width=8.5cm]{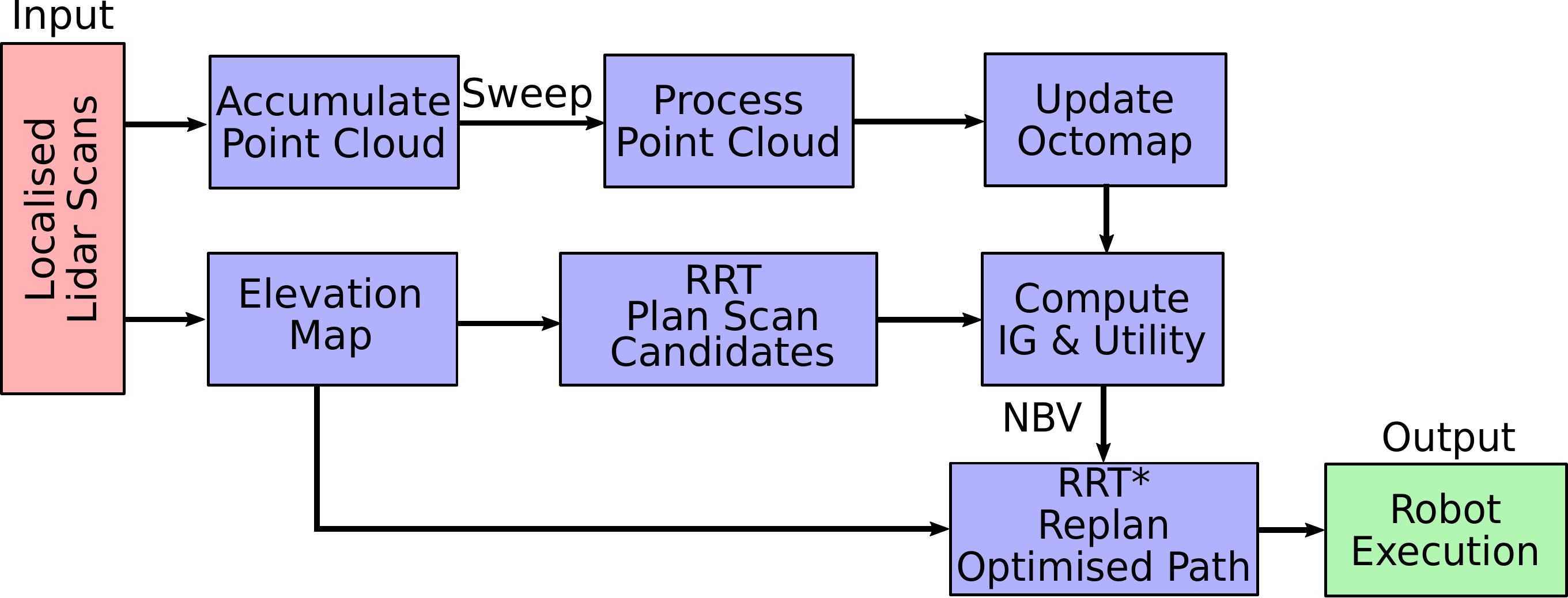}
    }
    \caption{Block diagram of the system architecture.}
    \label{fig:SystemArch}
    \vspace{-2em}
\end{figure}

In this section we detail the modules of our active mapping system.~\figref{fig:SystemArch} illustrates a 
block diagram of the system architecture. The system is based on an iterative pipeline. At the start of each iteration, the 
robot executes a scanning action while standing (further described in \secref{sec:Hardware}) to collect  
a sensor sweep. These measurements are incorporated into a map, a route to a new scan location is planned 
and the robot is requested to walk to the NBV for further exploration. This sequence is repeated, until a termination criterion (such as map completion) is met. 

The LiDAR measurements $\mathcal{L}_\mathcal{B}$ are sensed and then stored relative to the 
base frame $\{\mathcal{B}\}$. During a scanning action, $\mathcal{L}_\mathcal{B}$ is 
transformed into the map frame $\{\mathcal{M}\}$ based on the current pose of the 
robot $\bm{x}_\mathcal{M}\in\mathbb{R}^6$ and is accumulated into a larger point 
cloud. Our robot runs a localization system with little drift on the scale of our current experiments,
allowing us to assume that the pose $\bm{x}_\mathcal{M}$ is accurate. This is discussed further in~\secref{sec:Experiments}.

The accumulated point cloud is denoted \textit{sweep} $\mathcal{S}_\mathcal{M}$ 
in our system. $\mathcal{S}_\mathcal{M}$ is then downsampled for uniformity and 
filtered to remove outliers. Next, the system uses the processed \textit{sweep} $\mathcal{S}_\mathcal{M}$ as well as the pose of the robot $\bm{x}_\mathcal{M}$ to update the occupancy probabilities of voxels in its OctoMap. 

We also use the LiDAR measurements to generate an elevation map of the environment. The elevation map has a useful range of about $10$~m, allowing only local planning.
The path planning module evaluates terrain traversability subject to the elevation map and builds an RRT to generate a collection of scan candidates $\mathcal{C}$. A scan candidate $c\in\mathcal{C}$ is a pose where the robot could go to for the next scanning action.

We use a utility function $\mathcal{U}_{c}$ to determine the best scan candidate (NBV), $c_{best}$, from the set of candidates $\mathcal{C}$:
\begin{align}
\label{equ:utility_function}
\mathcal{U}_{c} = \mathcal{G}_{c}\times(1-\mathcal{P}_{c})\times(1-\mathcal{T}_{c}).
\end{align}
This function combines contributions from
\begin{itemize}
\item information gain $\mathcal{G}_{c}$: which measures the expected improvement of the model if given a sweep from that pose,
\item position cost $\mathcal{P}_{c}$: which penalises poses that have already been visited or are too close to the object,
\item traversal cost $\mathcal{T}_{c}$: which models the cost of travelling to a specific scan candidate pose.
\end{itemize}

These measures are discussed in the following sections. Finally, our system replans an optimised path using RRT*~\cite{Karaman2011} before the robot takes the next mapping action. 

\subsection{Volumetric Information (VI) Gain} 
\label{subsec:VIs}
Given a partial model of the object of interest, our system needs to determine the expected improvement in the model 
should a scan be made from a particular scan candidate pose. The approach is to trace a series of rays 
from a hypothetical pose and to estimate the expected information gain of observable voxels. 

Let $\mathcal{R}_c$ denote the set of rays cast by a scan candidate $c$. For each 
ray $r\in\mathcal{R}_c$, $\mathcal{V}_{r}$ is the set of voxels that the ray intersects with before reaching its endpoint. 
The information gain $\mathcal{G}_c$ at scan candidate $c$ is the sum of VIs, $\mathcal{I}$, in every voxel $v\in\mathcal{V}_{r}$ along each ray $r\in\mathcal{R}_c$:
\begin{align}
\label{equ:information_gain_sum}
\mathcal{G}_c = \sum_{\forall r\in \mathcal{R}_c}^{}\sum_{\forall v\in \mathcal{V}_r}^{}\mathcal{I}.
\end{align}

We implemented two formulations for $\mathcal{I}$ from Isler~\textit{et al.}~\cite{Isler2016}, namely
\textit{Occlusion Aware} $\mathcal{I}_{OA}$ and \textit{Rear Side Entropy} $\mathcal{I}_{RSE}$ which we summarise here.

Other proposed formulations are less relevant due to our sensor's long range and $360^\circ$ Field of View (FoV).

\subsubsection{Occlusion Aware}
This measure determines how effectively uncertainty will be reduced by scanning at a certain pose considering
voxel visibility.

Given the occupancy probability $P_o(v)$ of voxel $v$, the entropy of the voxel is obtained from: 
\begin{align}
\label{equ:entropy}
H(v) = \text{ - }P_o(v)\ln P_o(v) \text{ - } (1\text{ - }P_o(v))\ln (1\text{ - }P_o(v)). 
\end{align}

Then the \textit{Occlusion Aware} VI of voxel $v$, $\mathcal{I}_{OA}(v)$, is: 
\begin{align}
\label{equ:occlusion_aware}
\mathcal{I}_{OA}(v) = P_v(v)H(v), 
\end{align}
where $P_v(v)$ is the visibility probability of voxel $v$, which is computed as :
\begin{align}
\label{equ:visibility_proability}
P_v(v_n) = \prod_{i = 0}^{n-1} (1-P_o(v_i)), 
\end{align} 
where $v_n$ is the $n$-th voxel along the ray $r$ and $v_i$, $i = 0...n-1$, is a voxel 
that $r$ intersects before reaching $v_n$.

\subsubsection{Rear Side Entropy}
This measure is based on the \textit{Occlusion Aware} VI but focuses
on voxels at the rear of observed surfaces.
Rear Side Entropy is formulated as: 
\begin{align}
\label{equ:rear_side_entropy}
\mathcal{I}_{RSE}(v) = 
\begin{cases*}
\mathcal{I}_{OA}(v) & \text{$v$ is a \textit{Rear Side Voxel}, }\\
0 &\text{otherwise.}
\end{cases*}
\end{align}

The idea is that a \textit{Rear Side Voxel} is also likely to be occupied by the object. 
Focusing exploration on these voxels concentrates scans on the object rather than on surrounding 
free space. 
While these metrics proposed by~Isler~\textit{et al.}~\cite{Isler2016} are useful, their experimental validation
was limited to lab experiments with a stereo camera planning over a fixed set of poses.
We are motivated to develop a more complete field system which operates in a large scale industrial site.

Our system instead plans scan candidates progressively using an RRT which plans on the LiDAR elevation map. 
Our robot explores using a Velodyne LiDAR, which has a $360^\circ$ FoV 
horizontally and long sensing range, which is suitable for scanning large-scale objects or 
environments. 

In our experimental section (\secref{sec:Experiments}) we compare \textit{Rear Side Entropy} and \textit{Occlusion Aware} in field experiments. 

\subsection{Position and Traversal Cost}
\label{subsec:Costs}

In our path planning module (\secref{subsec:PathPlanning}), the RRT grows only within the traversable area of the elevation map, therefore the collection 
of scan candidates $\mathcal{C}$ does not contain invalid or 
unreachable poses. As a result,
the utility value $\mathcal{U}_c$ of each scan candidate is penalised 
based on the nature of ANYmal and the configuration of the LiDAR system. 

\subsubsection{Position Cost}

The position cost $\mathcal{P}_c$ is defined as: 
\begin{align}
\label{equ:position_cost}
\mathcal{P}_c = 
\begin{cases}
1-d_{thres}^{-1}\times d_c & d_{thres} \geq d_c \geq 0, \\
0 & d_c > d_{thres}, 
\end{cases}
\end{align}
where $d_c$ is the distance to an already visited scanning pose or the object itself, and $d_{thres}$ is a user defined threshold.
$\mathcal{P}_c$ is used to avoid rescanning a previously used region and maintains a reasonable distance between robot and object.

Using \textit{Occlusion Aware} VI, the system plans NBVs in regions where the robot can observe more void space. The main contribution to the information gain $\mathcal{G}_c$ is from void rays $r_{void}$ --- rays that do not hit any surfaces. Voxels $v_{void}\in\mathcal{V}_{r_{void}}$ are mainly unknown (occupancy probability $P_o(v_{void}) = 0.5$) and have high entropy. 

In our current system, $P_o(v)$ of a voxel $v\in\mathcal{V}_r$ is only updated when ray $r$ hits a surface, so $P_o(v_{void})$ does not change when observed by $r_{void}$. As a result for \textit{Occlusion Aware}, $\mathcal{I}_{OA}(v_{void})$ will not decrease, causing the robot to stop exploring. By applying the position cost, our system can also avoid visiting fully scanned areas. 
We plan to utilise $r_{void}$ to update voxel occupancy in the future version. 

In contrast, $r_{void}$ do not contribute to \textit{Rear Side Entropy} VI. 
Every scan decreases the entropy of observed voxels. 

In addition, the position cost $\mathcal{P}_c$ that applies to $c$ close to the object encourages candidates farther away, resulting our system observing a wider view. 
Conversely, if a scan candidate pose $c$ is farther away, less rays in $\mathcal{R}_c$ are able to observe the object, hence IG of this pose $\mathcal{G}_c$ would be lower compared to closer candidates. This discourages our system from selecting distant NBVs, ensuring a high resolution scan. 

Isler~\textit{et al.}~\cite{Isler2016} predefined a set of scan candidates in their system so that the distance of scan poses to the object surface was fixed. 
However, in our system, the distance between the robot and the scan surface is dynamic so that the robot can avoid obstacles in the environment. Furthermore, since the ANYmal operates on a 2.5D 
manifold, it is necessary for the quadruped to adjust the distance to the object surface so as to efficiently scan objects of different sizes. 
By combining IGs with a position cost, our system achieves a balance between coverage and resolution. 

\subsubsection{Traversal Cost}

The traversal cost $\mathcal{T}_c$ represents the difficulty for the robot to execute a certain path to candidate $c$ because of the roughness of terrain and the distance. 

Currently our approach classifies the elevation map discretely as either safe ($\mathcal{T}_c = 0$) or not traversable ($\mathcal{T}_c = 1$). 

In addition, a constant traversal cost penalises scan candidates that are behind the robot, because large turns are more difficult for the robot to execute. This policy also encourages the robot to explore forward rather than alternating direction. This makes the system more time and energy efficient. 

\subsection{Path Planning}
\label{subsec:PathPlanning}

The path planning module in our system consists of two phases, as indicated 
in~\figref{fig:SystemArch}. Both phases rely on an elevation grid map generated from LiDAR measurements of the environment. 
We used the approach of \cite{Fankhauser2018ral} to compute the slope and normal of each cell and in turn a measure of the traversability of the terrain.
The traversability is used to determine which states planned by the RRT and RRT* are valid and reachable. 

In the first phase, the RRT grows into the traversable area without a goal until a user-defined number of nodes have been generated. These nodes are scan candidates $\mathcal{C}$. 
We then compute the utility value $\mathcal{U}_c$ for each scan candidate $c\in\mathcal{C}$ independently and choose the NBV $c_{best}$ with the highest individual value. 
Following that, the second phase of our path planning module uses RRT* to replan the route to NBV, optimising travel distance. 

\subsection{Termination Condition}

In a model-free active mapping system, it is difficult to evaluate 
the completeness of reconstruction. 
We terminate operation using a user-defined threshold on the
utility value $u_{thres}$ after a planning sequence.

When the utility value of the NBV $\mathcal{U}_{c_{best}}$ falls below the threshold (\eqref{equ:termination}), 
no new scan candidate has satisfactory quality, and the active mapping procedure terminates. 
\begin{align}
\label{equ:termination}
\mathcal{U}_{c_{best}} < u_{thres} \quad \forall c \in \mathcal{C}.
\end{align}

\begin{figure}[!b]  
    \vspace{-1.5em}
    \centering
    \includegraphics[width=0.45\textwidth,,trim={0cm 0cm 0cm 5cm},clip]{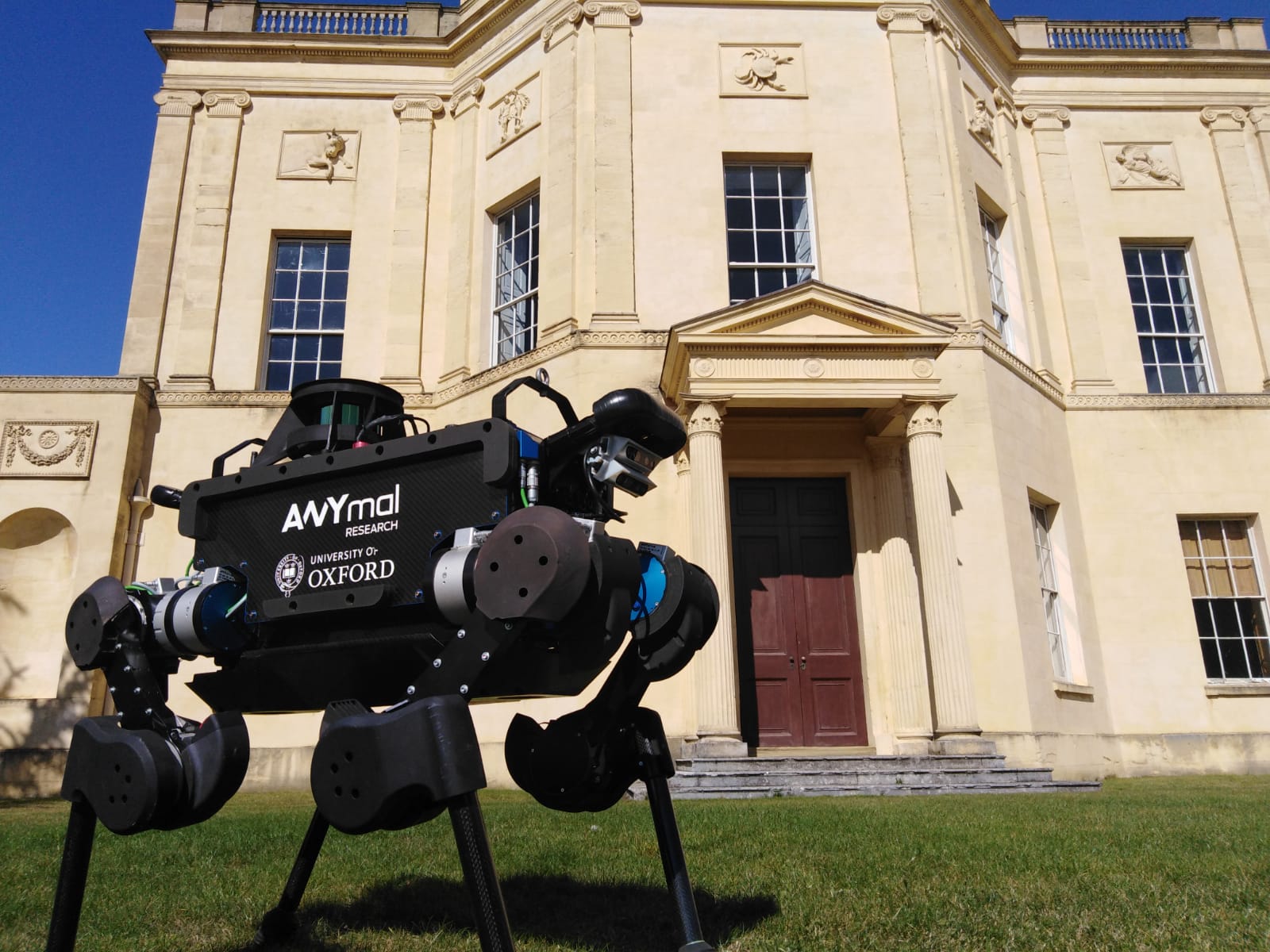}
    \caption{One of our experiments in~\secref{sec:Experiments} mapped this building facade at Green Templeton College (GTC), Oxford.}
    \label{fig:RealExperimentShowcase}
\end{figure}

\section{Experiments and Evaluation}
\label{sec:Experiments}
\begin{figure}[!t]
    
    \centering
    \resizebox{\columnwidth}{!}{%
        \includegraphics[width=0.6\textwidth,trim={7cm 9cm 7cm 9cm},clip]{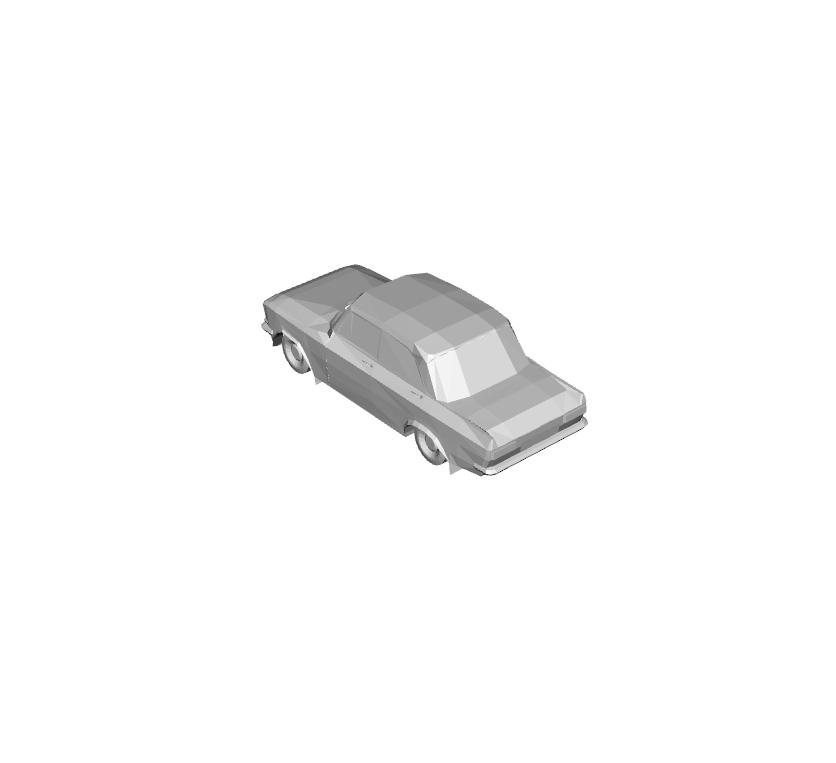}%
        \includegraphics[width=0.6\textwidth,trim={4cm 6cm 4cm 4cm},clip]{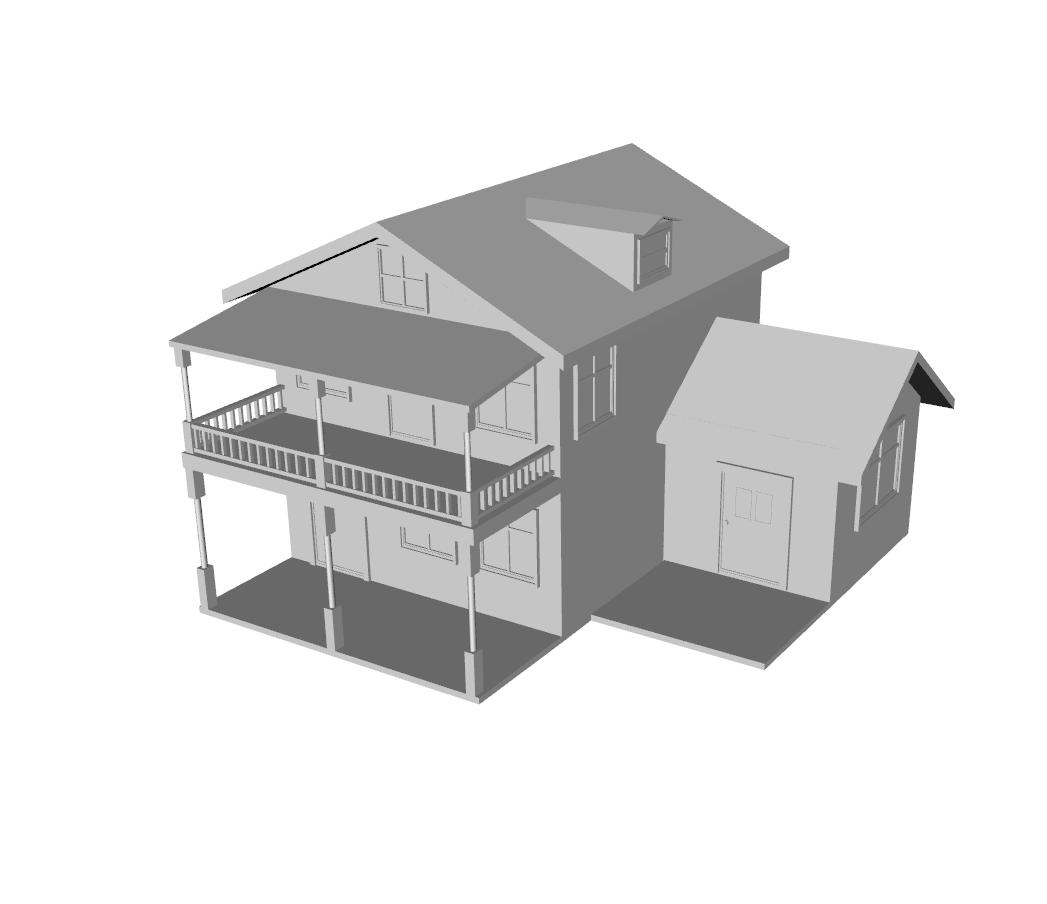}
    }
    \resizebox{\columnwidth}{!}{%
        \includegraphics[width=0.4\textwidth,trim={2cm 1cm 2cm 2cm},clip]{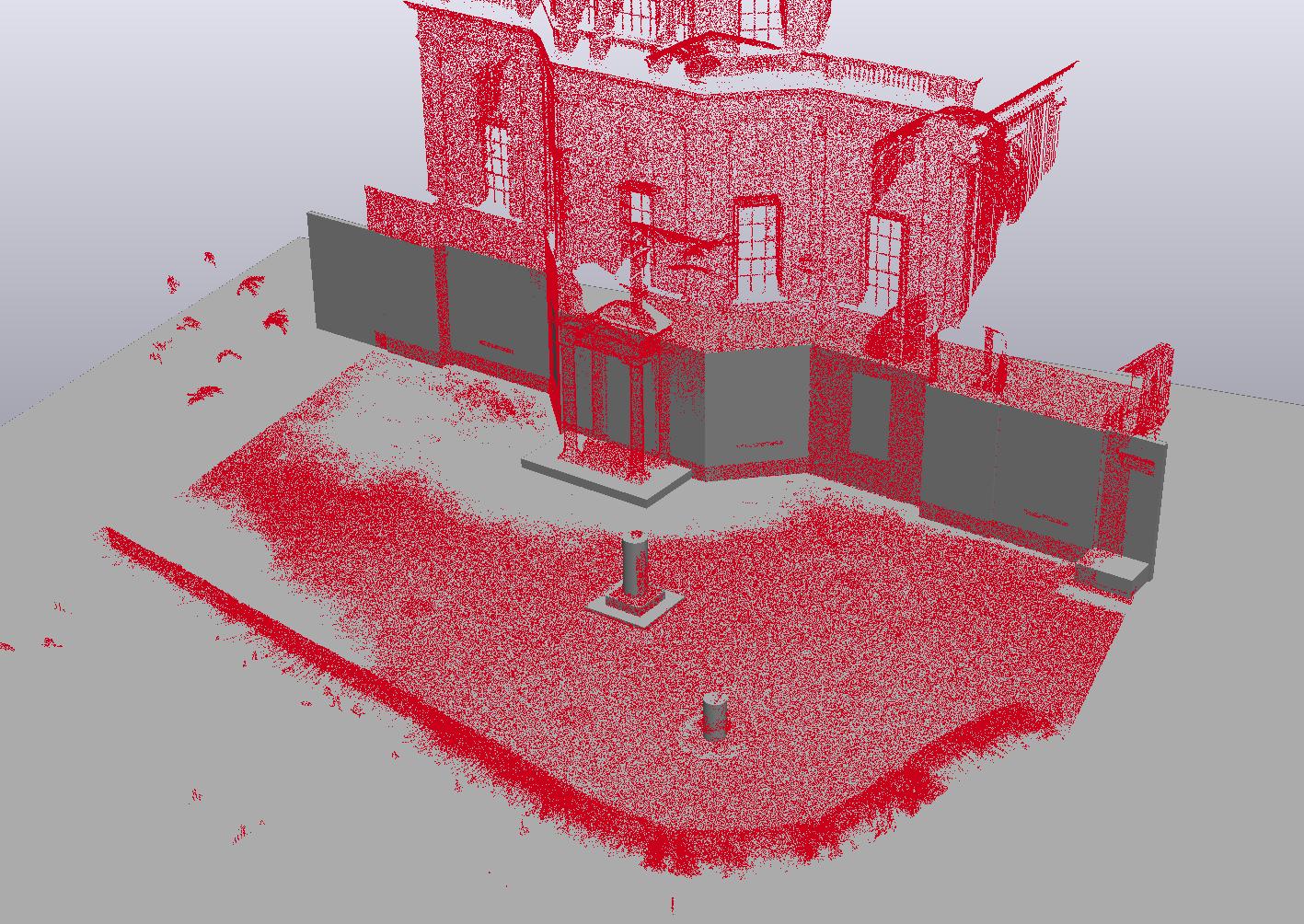}
        \includegraphics[width=0.4\textwidth,trim={2cm 4cm 2cm 3.4cm},clip]{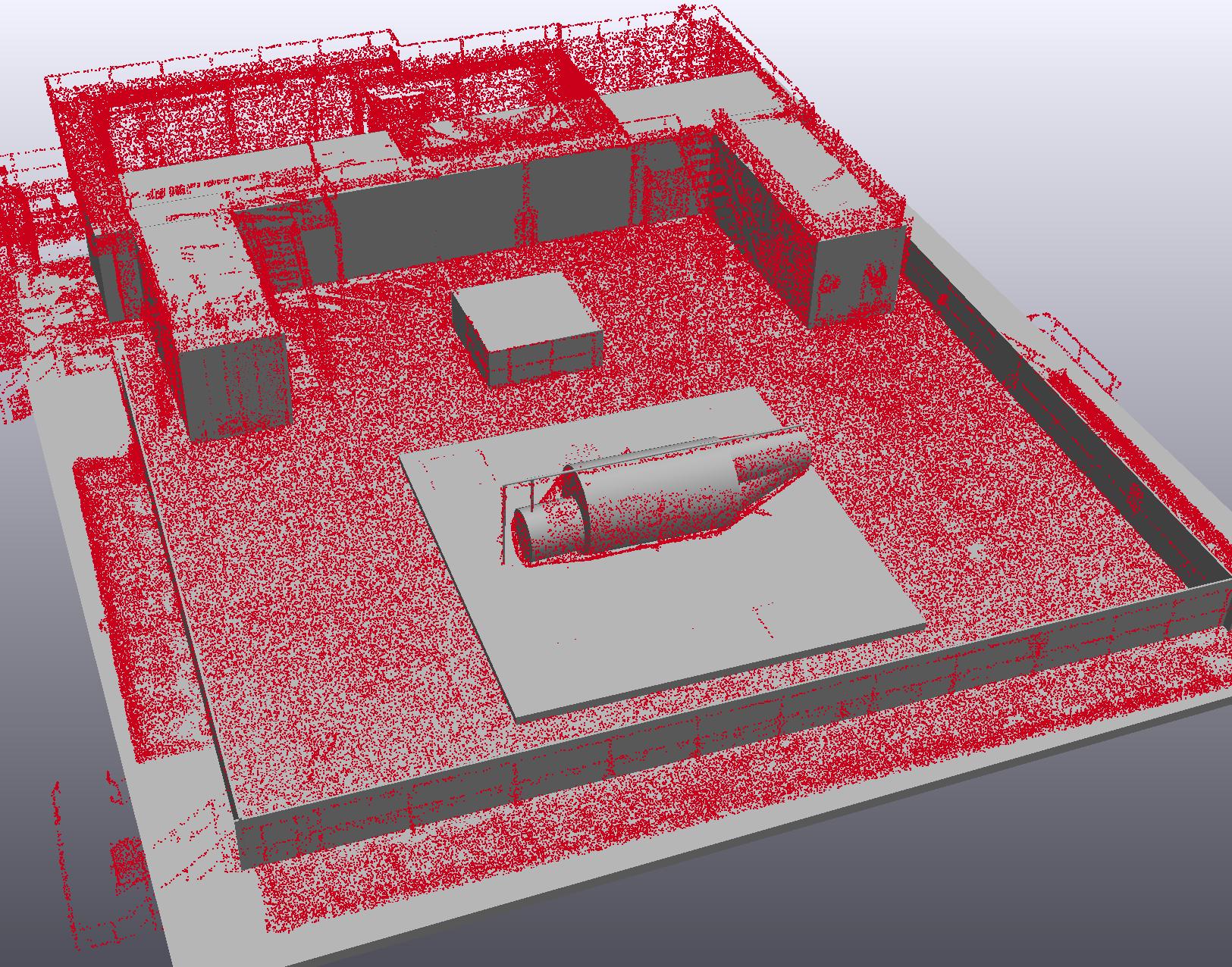}
    }
    \caption{3D models of a car and house (top) and of our test sites (bottom) are used to evaluate our system. We created the 3D models using reconstructions made using a survey-grade LiDAR (in red).}
    \label{fig:SimulationModels}
    \vspace{-2em}
\end{figure}

To demonstrate our system's functionality and to test the VI formulations, we carried out evaluations of increasing complexity---with the simple virtual models in \figref{fig:SimulationModels} and Gazebo reconstructions of our envisaged test locations to test our system's ability to avoid collisions. Finally, we deployed our system on the real ANYmal robot in these environments. The results are detailed in the following sections. 

The real experiments involved scanning a building facade at Green Templeton College (GTC) 
($10\times35\times4$~m$^3$) in Oxford (\figref{fig:RealExperimentShowcase}) and a mock-up helicopter on the oil 
rig training site at the Fire Service College (FSC) ($3\times8\times3$~m$^3$) in Gloucestershire (\figref{fig:FscHelicopterDeck}). 

In these experiments, we used a LIDAR localization system running on the robot's navigation PC. The system registered LiDAR clouds against a prior point cloud map using Iterative Closest Point (ICP)~\cite{pomerleau13ar} seeded with legged odometry. At the scale of our experiments, a deformable map representation was not needed.

\subsection{Hardware}
\label{sec:Hardware}
\subsubsection{Platform}
The robot platform employed in this work is an ANYbotics ANYmal (version B)~\cite{hutter2016anymal}. 
The robot has 12 actuated joints, as well as the 6 DoF floating base link. It is capable of trotting at the maximum speed of $1.0$~m/s and traversing complex terrain, e.g. stairs, kerbs and ramps. 

\subsubsection{Sensor}
The primary sensor of our system is a Velodyne VLP-16 LiDAR which has 
16 laser beams spread across a $\pm15.0^\circ$ vertical FoV 
and measures ranges with the accuracy of $\pm3$~cm across full $360^\circ$ horizontally.  

Utilising the robot's wide range of motion, we designed a scanning action to roll the base from $40^\circ$ to $-40^\circ$ (while standing). The action improves the vertical FoV of our system to $\pm55^\circ$ and allows mapping objects much taller than the robot. Using this action, our system collects individual LiDAR sweeps. 

\subsection{Simulated Experiments}

We conducted experiments in simulation to map models of a car and a house (\figref{fig:SimulationModels} (top)). 
We then used a Leica BLK360 laser scanner to create accurate reconstructions of our two
test sites, the facade of a building and a helicopter deck ((\figref{fig:SimulationModels} (bottom)). We modelled the major surfaces of these test sites to create Gazebo simulations of the test sites.

In these experiments, the approximate location and size of the object of interest are 
known, which aids segmentation from the accumulated sweep. This 
informs our system about where the OctoMap should be constructed 
and the VIs be computed. We chose a $5$~cm resolution for our OctoMap octree, which suits the resolution of the Velodyne LiDAR.

For path planning and NBV selection, our system grows an RRT up to $150$ nodes, every iteration, 
within a $12\times12$~m$^2$ elevation map centred around the robot. 
This allows the robot to plan and conduct mapping actions around 
the object. 

To quantify the mapping results, we exploit different criteria including point 
cloud coverage ($c_p$), travel distance ($d_t$) and number of scan actions 
($n_s$). In addition, we compute the overall task time ($t_{all}$) as well as 
the average time per-scan spent computing information gains and determining 
the NBVs ($t_{nbv}$) to evaluate the system's online feasibility.

To compute point cloud coverage, we aligned the accumulated point cloud with the ground truth and determined the points in the accumulated cloud within $4.3$~cm of the nearest point in the ground truth, approximately the 
distance between the centre of our OctoMap voxel and its vertex ($\frac{\sqrt{3}}{2}\times5$~cm). These points are classified as observed. 

Point cloud coverage $c_p$ is then defined as: 
\begin{align}
c_p &= \frac{N_{O}}{N_{GT}}
\end{align}
where $N_{GT}$ and $N_{O}$ are the total number of points in the ground truth model and the number of points observed in the model so far, respectively. 

\begin{figure}
    \centering
    \includegraphics[width=0.5\textwidth,trim={2cm 0cm 0cm 1cm},clip]{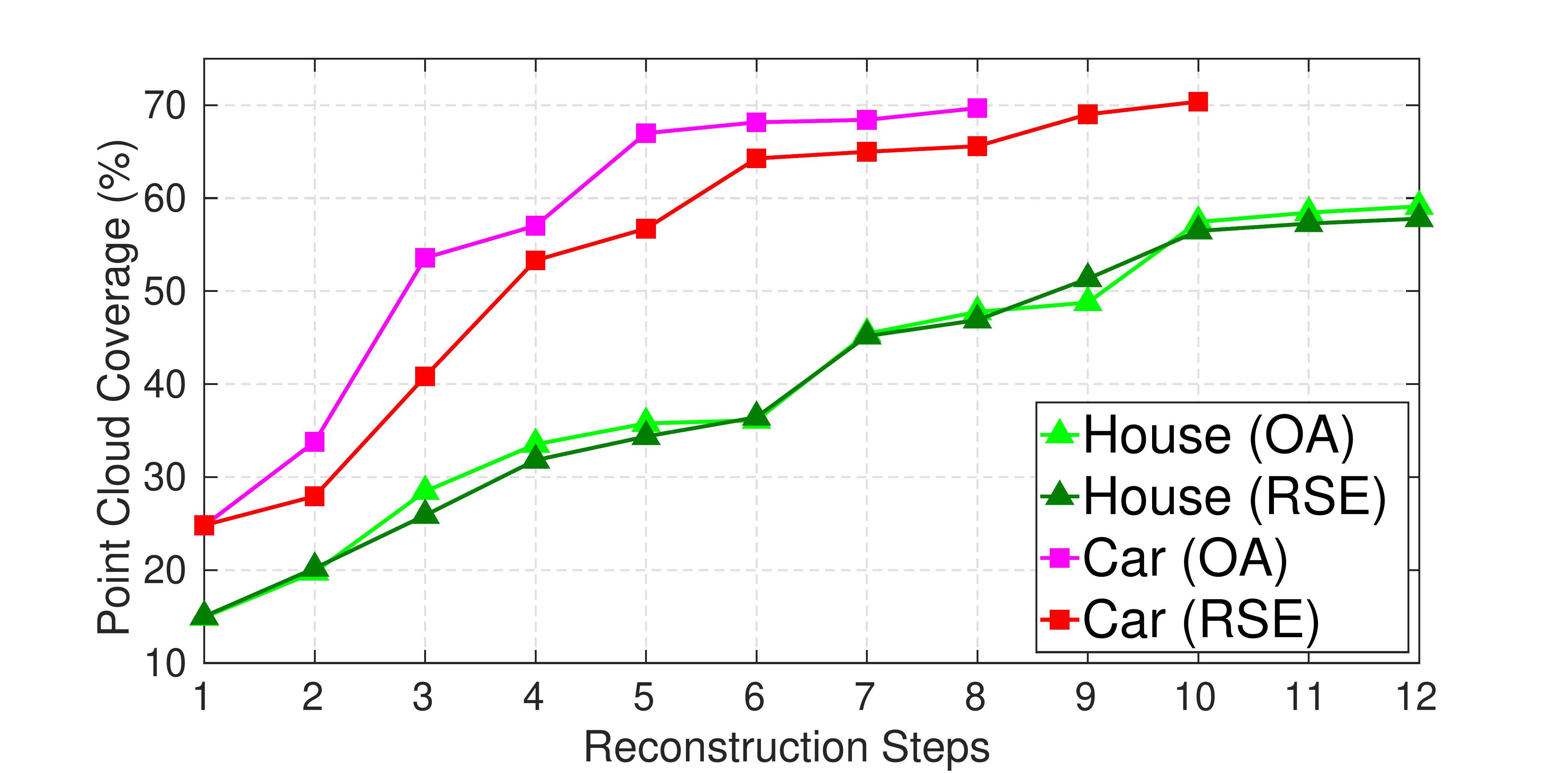}
    \vspace{-2em}
    \caption{Point cloud coverage per step for the car and house models.}
    \label{fig:StepCoverage}
\end{figure}

As summarised in~\tabref{table:SimulationComparison}, the point cloud coverage gained with \textit{Occlusion Aware} IG is slightly higher than that with \textit{Rear Side Entropy} IG. This can also be seen in~\figref{fig:StepCoverage}, which demonstrates the point cloud coverage per scan. The maximum coverage never reaches $100\%$ in our system as the top surfaces of objects are higher than the robot and cannot be observed from the ground, as shown in \figref{fig:HelicopterCoverage}. 

While there is on average an 8.5\% reduction in travel distance when our system employs \textit{Occlusion Aware} compared to \textit{Rear Side Entropy}, our system is subject to the random scan candidate placement by RRT. Hence the performance difference between two VIs in simulation so far is not significant enough for us to make a conclusive decision on which is the better formulation. Both approaches allowed our system to accomplish the mapping task. In both cases, the travel distance, the overall run time and the NBV computation time are all feasible for real experiments. 

\begin{figure}[!t]
    \centering
    \includegraphics[width=0.4\textwidth,trim={0cm 0cm 0cm 3cm},clip]{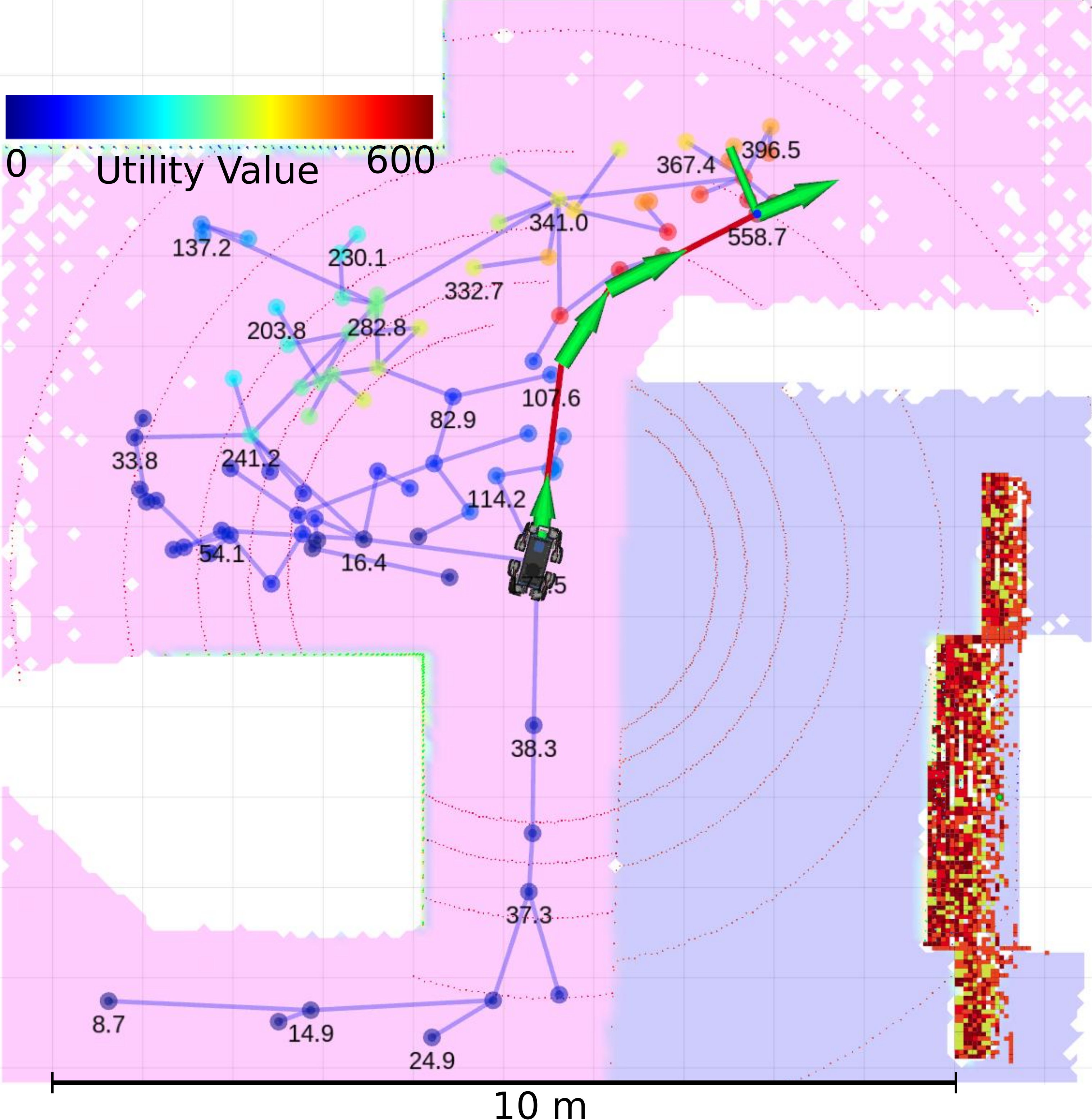}
    \caption{Illustration of the proposed system mapping the simulated model of the helicopter/oil rig site. The system grows an RRT around the current pose of the robot in the traversable area. The utility metric is computed at the tree nodes. The node with the maximum utility is selected as the NBV. Finally, using the RRT* algorithm, the path from the current pose to the NBV is replanned.}
    \label{fig:RealSimExamples}
    \vspace{-1.5em}
\end{figure}

\begin{table}[t]
    \centering
    \resizebox{\columnwidth}{!}{%
    \begin{tabular}{|lc|ccccc|}
        \hline \cellcolor{orange!25}& \cellcolor{orange!25}& \multicolumn{5}{c|}{\cellcolor{orange!25}Experiments Evaluation} \\  
        \cellcolor{orange!25}Object& \cellcolor{orange!25}IG& \cellcolor{orange!25}$c_p$ (\%) & \cellcolor{orange!25}$d_t$~(m)  & \cellcolor{orange!25}$n_s$ & \cellcolor{orange!25}$t_{all}$~(mm:ss) & \cellcolor{orange!25}$t_{nbv}$~(sec)  \\ 
        \hline \hline 
        \multirow{2}{*}{Car} & OA & 69.69 & 35.39 & 8 & 08:02 & 2.70 \\ 
                             & RSE & 69.01 & 40.18 & 10 & 10:21 & 2.46 \\ 
        \hline 
        \multirow{2}{*}{House} & OA & 59.41 & 42.89 & 12 & 10:06 & 3.78 \\ 
                               & RSE & 58.98 & 44.64 & 12 & 11:17 & 3.65 \\ 
        \hline
        \multirow{2}{*}{Facade} & OA & 95.11 & 33.98 & 9 & 07:44 & 17.82 \\ 
                                & RSE & 94.24 & 37.82 & 9 & 08:21 & 19.69 \\ 
        \hline 
        \multirow{2}{*}{Helicopter} & OA & 83.76 & 41.56 & 12 & 12:26 & 5.45 \\ 
                                    & RSE & 83.01 & 43.61 & 13 & 13:06 & 5.20 \\ 
        \hline
    \end{tabular}
    }
    \caption{Comparison between two VI measures in simulation environments (OA - \textit{Occlusion Aware}; RSE - \textit{Rear Side Entropy}).}
    \label{table:SimulationComparison}
    \vspace{-2em}
\end{table}

\subsection{Real-World Experiments}
Based on the simulated results in the previous section, we used the~\textit{Occlusion Aware} VI gain metric in our real experiments.

\tabref{table:RealExperiments} summarises the evaluation of the reconstruction results in both experiments. 
In these, because our elevation map is partly corrupted by odometry noise (see attached video) and the LiDAR sensor is just $70$~cm from the ground, we can only plan in a $7\times7$~m$^2$ area around the robot. 
We therefore decreased the number of RRT scan candidates from $150$ to $75$, consequently 
decreasing the computation time of determining the NBV $t_{nbv}$. 

Comparing \tabref{table:RealExperiments} with \tabref{table:SimulationComparison}, one can see that the computation times for the real experiments at facade and helicopter locations are on average half of the time taken in simulation.

Our approach allows the robot to avoid the mobility hazards for the helicopter experiment in~\figref{fig:FscHelicopterDeck}: stairwells, open edges on the deck and a skirt around the helicopter. 

\begin{figure}[t!]
    \centering
    \includegraphics[width=0.4\textwidth,trim={0cm 0.5cm 0cm 0.5cm},clip]{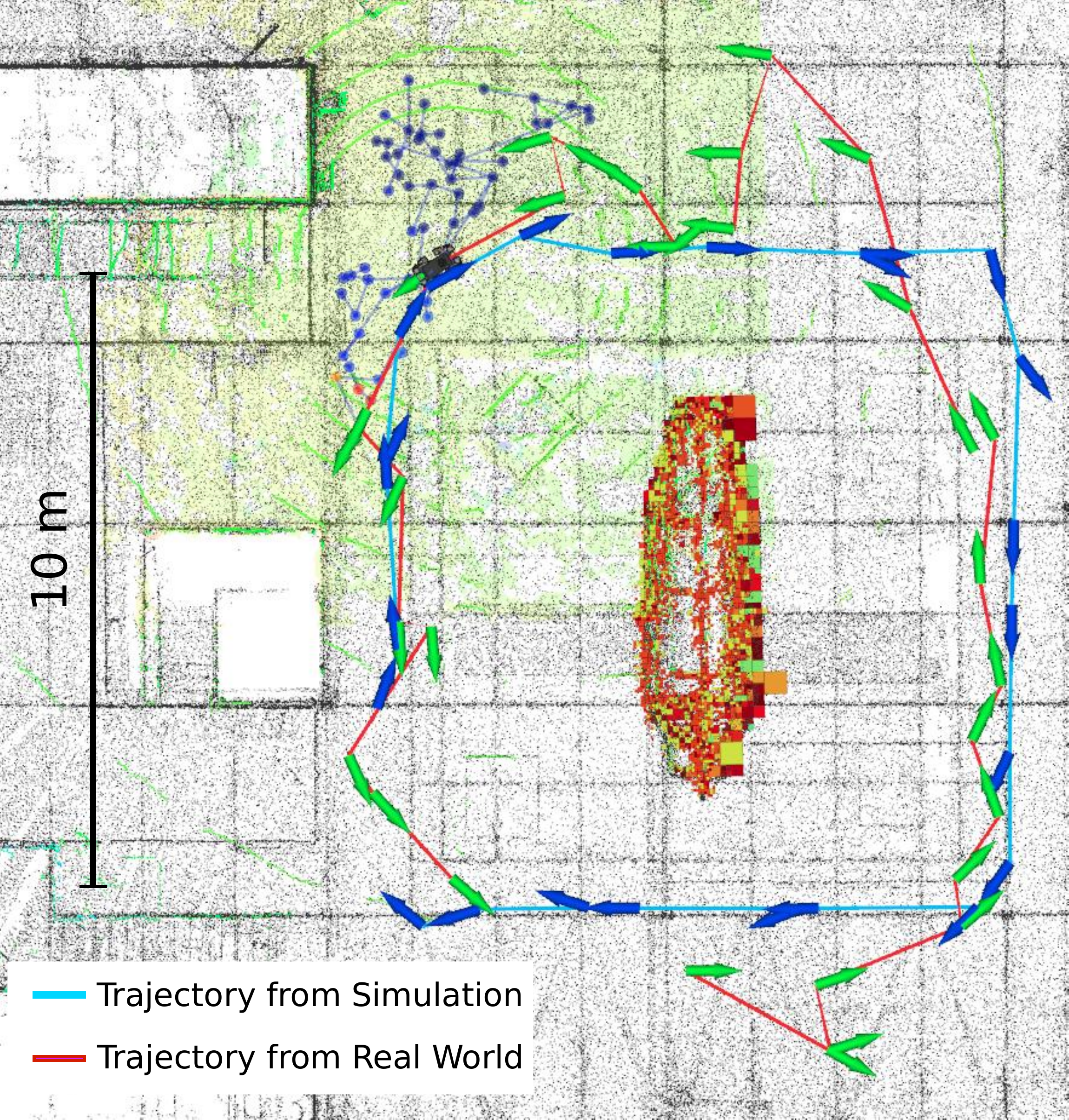}
    \caption{Example of our system mapping the real helicopter. The route the robot took in the real experiment is shown in red with the reconstruction of the helicopter (as an octree) shown in the centre. Also illustrated, in blue, is the route taken by our method running in simulated model, as a comparison.}
    \label{fig:HelicopterExample}
    \vspace{-1em}
\end{figure}

\begin{table}
    \centering
    \resizebox{\columnwidth}{!}{%
    \begin{tabular}{|l|ccccc|}
        \hline \cellcolor{orange!25}& \multicolumn{5}{c|}{\cellcolor{orange!25}Experiments evaluation} \\  
        \cellcolor{orange!25}Object& \cellcolor{orange!25}$c_p$ (\%) & \cellcolor{orange!25}$d_t$~(m)  & \cellcolor{orange!25}$n_s$ & \cellcolor{orange!25}$t_{all}$~(mm:ss) & \cellcolor{orange!25}$t_{nbv}$~(sec)  \\ 
        \hline \hline 
        Facade & 88.06 & 37.61 & 15 & 20:56 & 9.82 \\ 
        \hline 
        Helicopter & 78.60 & 49.19 & 26 & 35:25 & 2.00 \\ 
        \hline
    \end{tabular}
    }
    \caption{Results for our system in the real-world experiments.}
    \label{table:RealExperiments}
    \vspace{-1.5em}
\end{table}

\begin{figure}[t!]
    \centering
    \includegraphics[width=0.35\textwidth,trim={0cm 2cm 0cm 2cm},clip]{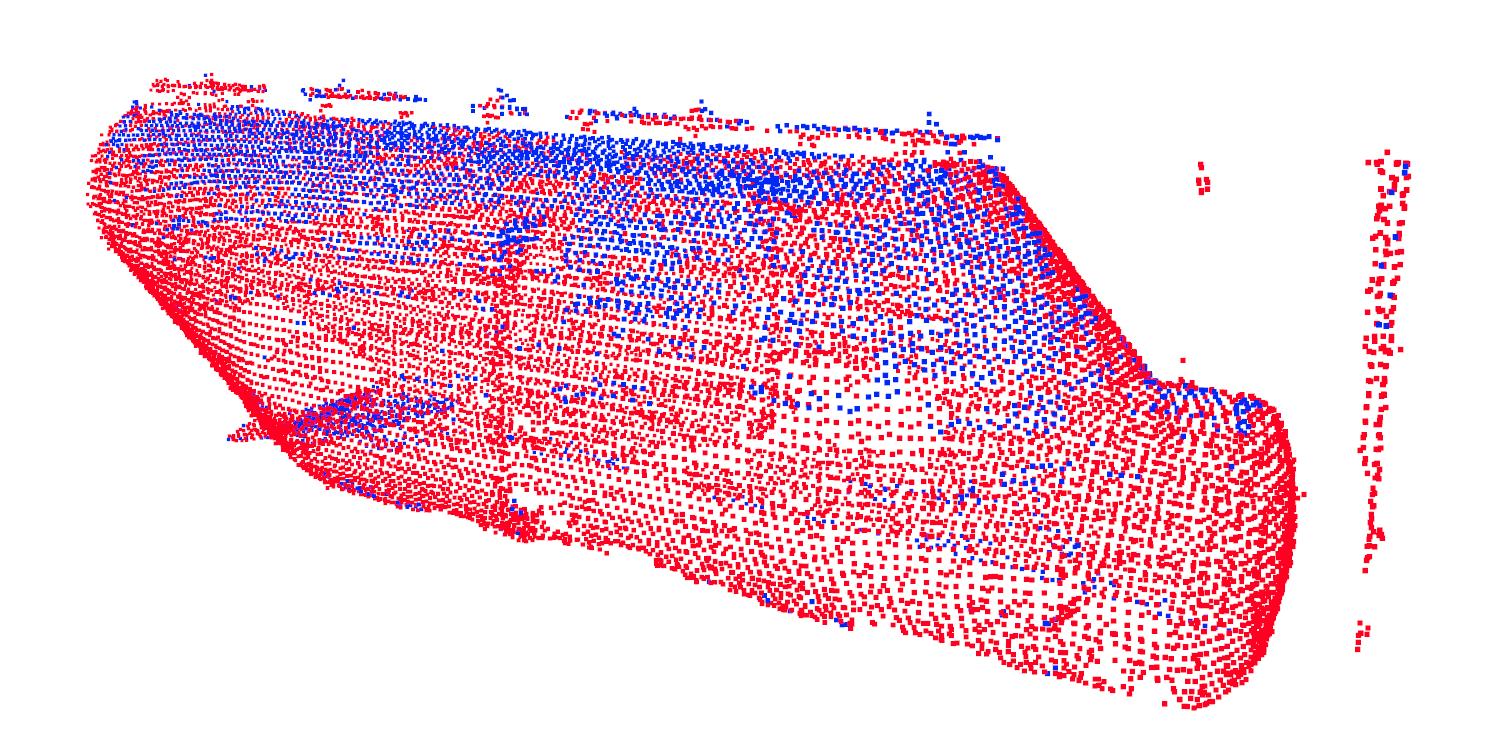}
    \includegraphics[width=0.35\textwidth,trim={1cm 3cm 2cm 4cm},clip]{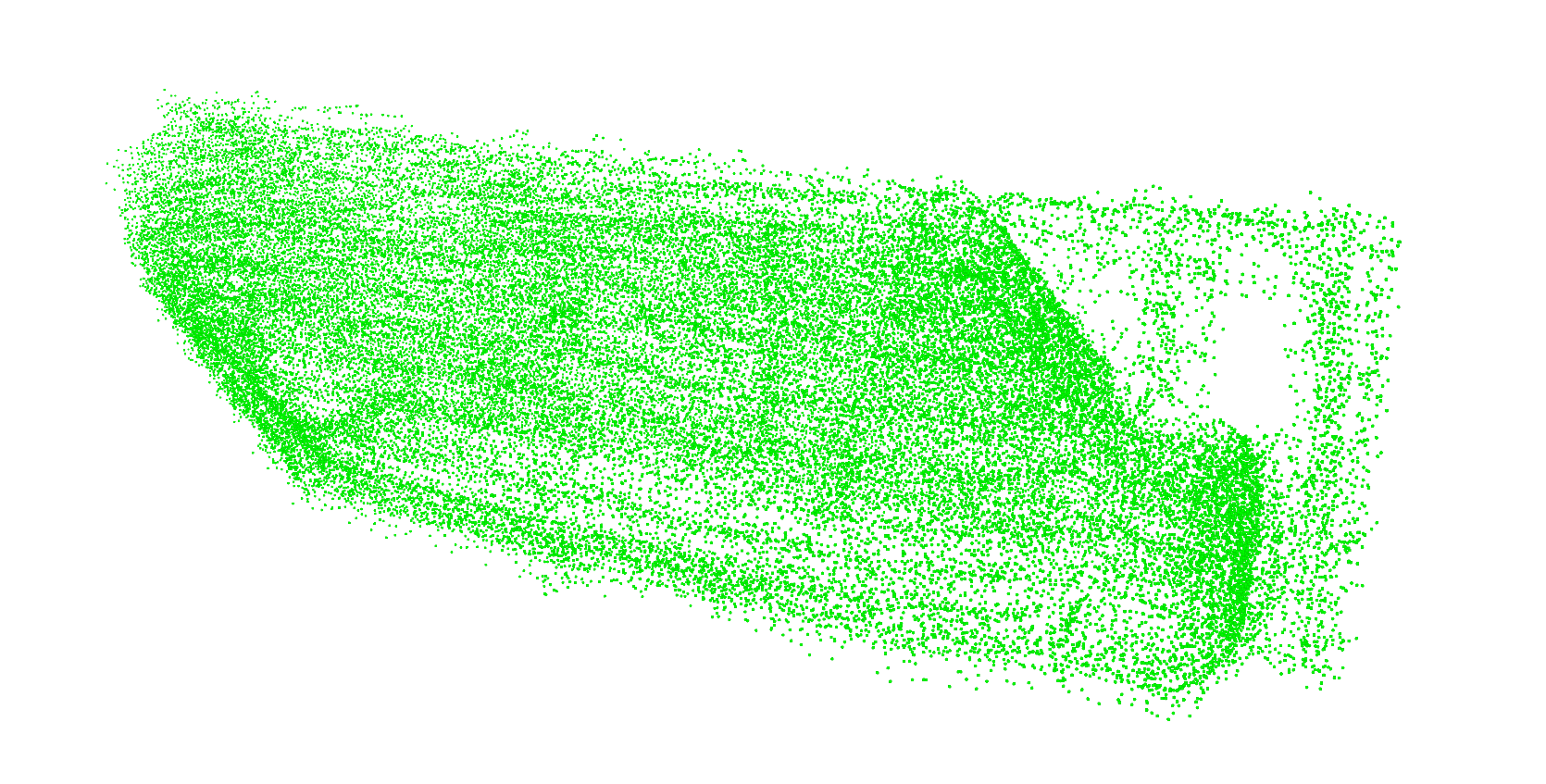}
    \caption{Top: Point cloud coverage in the real helicopter experiment. Red cloud indicates the observed area and blue represents the unobserved part of the helicopter. Bottom: Our system succeeded in reconstructing an accurate model of the helicopter body.}
    \label{fig:HelicopterCoverage}
    \vspace{-1em}
\end{figure}

\figref{fig:HelicopterExample} demonstrates a comparison between the robot trajectories in the real FSC helicopter experiment (counter-clockwise) and in simulation (clockwise). 
The paths taken by our system in both scenarios are similar, demonstrating the practicality of our system in real scenarios. However, as noted before, because of more unknown areas existing within the elevation map, the system had to plan around void cells. 

\figref{fig:HelicopterCoverage} demonstrates the success of our system in reconstructing the helicopter body (in green), compared to the ground truth (in red).  
Due to the limitation in the elevation map and in turn our path planning module,
our system planned more scans than in simulation.

In the helicopter scenario, the number of scans required more than doubled and as a result the run time almost tripled. A major part of that difference is due to the time spent by the robot operator judging if planned paths were safe.

\section{Conclusion and Future Work}
\label{sec:FutureWork}
In this work we presented a model-free active mapping system using a volumetric representation. 
The system allows a quadruped robot to explore and reconstruct both small and large scale objects, in particular industrial assets, with few assumptions about the test environment and requiring only high level human supervision. 
We tested our system in fully realistic scenarios and our approach allowed the robot to accomplish mapping missions in a complicated environment, creating accurate reconstructions online. 

In the future, we plan to improve the quality of elevation map and to incorporate full traversability estimation to allow our robot to navigate over rough terrain, such as kerbs and ramps, so as to fully utilise the dynamics of a legged robot. 

In addition, we plan to base localisation on a pose-graph SLAM system~\cite{ramezani2020online} so that our active mapping approach can explore larger environments, with the benefit of loop closure. Hence, we would like to modify the reconstruction to be deformable, as in the manner proposed by Ho~\textit{et al.}~\cite{Ho2018}. 

\balance
\bibliographystyle{IEEEtran}
\bibliography{library}

\end{document}